# Deep learning applied to EEG data with different montages using spatial attention

Dung Truong*
SCCN, INC, UCSD, La Jolla CA, USA
dutruong@ucsd.edu
https://orcid.org/0000-0003-4540-3551

Muhammad Abdullah Khalid*
SCCN, INC, UCSD, La Jolla CA, USA
mkhalid.bee18seecs@seecs.edu.pk

Arnaud Delorme
SCCN, INC, UCSD, La Jolla CA, USA
CerCo CNRS, Paul Sabatier University, Toulouse, France
arnodelorme@gmail.com
https://orcid.org/0000-0002-0799-3557

*Abstract—* The ability of Deep Learning to process and extract relevant information in complex brain dynamics from raw EEG data has been demonstrated in various recent works. Deep learning models, however, have also been shown to perform best on large corpora of data. When processing EEG, a natural approach is to combine EEG datasets from different experiments to train large deep-learning models. However, most EEG experiments use custom channel montages, requiring the data to be transformed into a common space. Previous methods have used the raw EEG signal to extract features of interest and focused on using a common feature space across EEG datasets. While this is a sensible approach, it underexploits the potential richness of EEG raw data. Here, we explore using spatial attention applied to EEG electrode coordinates to perform channel harmonization of raw EEG data, allowing us to train deep learning on EEG data using different montages. We test this model on a gender classification task. We first show that spatial attention increases model performance. Then, we show that a deep learning model trained on data using different channel montages performs significantly better than deep learning models trained on fixed 23- and 128-channel data montages.

## I. INTRODUCTION

Deep learning is commonly used for the end-to-end classification of electrophysiology (EEG) data or as a feature transformer for EEG data [1, 4]. However, training deep learning models requires a large amount of data, and this is often a limitation with most EEG datasets where only a handful of subjects are available. A solution to the scarcity of large EEG datasets is to combine datasets from different sources. However, a challenge to this approach is the harmonization of data with different channel montages. Different EEG caps are being used in different EEG experiments, so the number and locations of channels/electrodes are rarely the same.

A straightforward solution would be to subsample the channel space, only selecting common channels across experiments when it is even possible. This approach, however, potentially underutilized available data. Previous works have been solving this problem by transforming raw EEG signals into a common feature space, such as the topography distribution of spectral power [4] or the spectrogram of single-channel EEG [16], before using deep neural networks. These solutions are nevertheless limiting since only spectral information of specific frequency bands is being considered.

Recent works are now showing that deep models trained on raw EEG data might learn statistical properties of the data well beyond the commonly used EEG frequency bands [14] and that deep learning models trained on raw EEG outperform models trained on spectral features [12]. Spectral approaches limit the data-driven capacity of deep learning models, which could potentially extract more information from the raw EEG data. Thus, researchers should explore channel harmonization methods for the training of deep learning models on raw EEG data from datasets with different montages.

In recent works, a mechanism called spatial attention was implemented to train deep models to correlate EEG/MEG with speech data [10]. In this method, deep learning (DL) models take into account the relative position of EEG electrodes on the scalp – information that is not available when DL models use as input 2-D *channel* x *time* EEG segment time series. This procedure consists of mapping channel locations from the coordinate space into a 2-D Fourier space where spatial dimensions are defined by channel 2-D locations proximity, then applying the attention mechanism to the frequency-transformed channels to map the input channels to a fixed number of output channels. While Défossez and collaborators [10] applied this mechanism to leverage information pertaining to channel spatial distribution, this could also be used to map different channel montages into a common output channel space.

In this report, we explore the application of the aforementioned spatial attention mechanism to the problem of training a single deep-learning model on data with different input channel montages. We used a large EEG resting state dataset from more than a thousand juvenile (5-22 years) participants, collected and made publicly available by the Child Mind Institute Healthy Brain Network project. We performed gender classification using a simple convolution deep neural model [12], to which we added spatial attention. To instigate the usefulness of using the spatial attention mechanism for channel harmonization, we performed data subsampling to obtain a subset of the data with two different

---
* authors contributed equally

channel montages, one with 128 channels and one with 23 channels.

## II. METHODS

**EEG recordings**. High-density EEG data were recorded in a sound-shielded room at a sampling rate of 500 Hz with a bandpass of 0.1 to 100 Hz, using a 128-channel EEG geodesic hydrogel system by Electrical Geodesics Inc. (EGI) [13]. The data are publicly available for download at http://fcon_1000.projects.nitrc.org/indi/cmi_healthy_brain_network. We only considered the resting data files. These were 6 minutes in length and were composed of successive 20-s to 40-s periods of eyes open, and eyes closed rest, respectively.

**Raw data preprocessing**. Although deep learning may be applied to raw EEG data without any preprocessing [1], we minimally preprocessed the data [2] using the EEGLAB v2023 software package [5] running on MATLAB 2022b. We used only eye-closed data segments (~170s per subject), ignoring the first and last 3 seconds of each eye-closed period (resulting in five periods of 34 seconds). We removed the mean for each data epoch from each channel, down-sampled the data to 128 Hz, and subsequently band-pass filtered the data between 0.25–25 Hz (FIR filter of order 6601; 0.125 Hz and 25.125 Hz cutoff frequencies (-6 dB); zero phase; non causal). Data were re-referenced to the averaged mastoids and cleaned using Artifact Subspace Reconstruction EEGLAB plug-in *clean_rawdata* (v2.3) [6], an automated method that removes artifact-dominated channels (parameters used were 5 for *FlatLineCriterion*, 0.7 for *ChannelCriterion*, and 4 for *LineNoiseCriterion*). Removed channels were then interpolated using 3-D spline interpolation (EEGLAB *interp.m* function). No bad portions of data were removed. Finally, we segmented eye-closed data periods into non-overlapping 2-s windows: each preprocessed 2-s epoch was used as a sample for our final dataset. Each subject provided about 81 2-s samples (mean 80.8 ± 3.32). The 128-Hz down-sampling procedure and 2-s window length extraction were identical to those used in Van Putten et al. [2]. No bad epochs were removed, and no further preprocessing was performed.

**Deep learning model.** VGG-16 was originally designed to categorize 15 million images of dimension 256x256 with 3 color channels [15]. In a previous publication, we successfully applied VGG-16 to EEG spectral and raw data classification after removing some of the layers to account for the reduced dimensionality of EEG data [4, 12]. We decreased the model complexity by reducing the number of convolutional layers, omitting layers 19-32 of VGG-16. We also divided the number of filters and hidden units in the convolutional and FC layers by 4 to reflect our lower number of training samples. To allow the model to take raw EEG data as input, we decreased the number of input channels in the first convolutional layer from 3 to 1. The rest of the network remained unchanged. In total, this model, called R-VGG, contained 7,452,850 trainable parameters. We chose this model to benchmark attention since it is simple and leads to good performance on the task of gender classification of EEG data [12].

| Layer | Filter size | # of filters/hidden units |
|---|---|---|
| Convolutional | 3x3 | 16 |
| Convolutional | 3x3 | 16 |
| MaxPooling | | |
| Convolutional | 3x3 | 32 |
| Convolutional | 3x3 | 32 |
| MaxPooling | | |
| Convolutional | 3x3 | 64 |
| Convolutional | 3x3 | 64 |
| Convolutional | 3x3 | 64 |
| MaxPooling | | |
| Fully connected | | 1024 |
| Dropout (50%) | | |
| Fully connected | | 1024 |
| Dropout (50%) | | |
| Fully connected | | 2 |
| Softmax | | |

**Table 1. R-VGG configurations**. The ReLU activation function is not shown for brevity. All convolutional layers have stride 1 and padding 1. All pooling layers have window size 2x2, stride 2, and no padding.

**Attention mechanism.** Attention is a sequence-to-sequence mechanism in which each element of the output sequence is a weighted sum of the elements of the input sequence. The goal of the attention mechanism can be interpreted as finding the relevance of the input elements to the output element. The general attention mechanism works by having the query, key, and value matrices. The computation steps start with combining the query and key, which can be implemented as a dot product or a linear transform:

$$a(Q, K) = Q.K^T \quad (1)$$

We then score the linear transformation of the input using a scoring function, most often softmax:

$$\alpha(Q, K) = softmax(a(Q, K)) \quad (2)$$

Then finally, the attention output will be the weighted sum of the elements of value matrix *V* given the attention score $\alpha(Q, K)$:

$$attn(Q, K, V) = \alpha(Q, K) V \quad (3)$$

This algorithm mostly finds applications in Natural Language Processing. However, its use has been expanded to Computer Vision and EEG Signal Processing. Su and collaborators [9] proposed a spatio-temporal attention network (STANet) in which they calculate spatial and temporal EEG feature representation. Zhang et al. [8] use a "Learnable Spatial Mapping Module" for projecting and averaging the EEG channels to a new domain.

**Spatial attention.** Défossez and collaborators [10] applied the attention mechanism to multi-channel electrophysiological data by first parameterizing the electrode coordinates by a 2D frequency space. The 3D channel positions are converted to a 2D plane using the MNE-Python function *find_layout*. These 2D positions are then normalized to a range of 0 to 1 (Fig. 1).

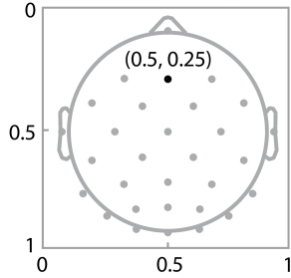

**Figure 1.** Encoding of electrode location in normalized 2-D space. The (x,y) coordinate of the channel mark in black is shown. Only a subset of the 128 channels are shown for clarity.

For an output channel $j$, the equation below computes the score for each input channel location $(x, y)$ where $Re$ and $Im$ are the real and the imaginary part of the elements of $z_j$ respectively:

$$a_j(x,y) = \sum_{k=1}^{K}\sum_{l=1}^{K} Re(z_j^{(k,l)})cos(2\pi(kx+ly)) + Im(z_j^{(k,l)})sin(2\pi(kx+ly)) \quad (4)$$

One interpretation of this equation is that it is letting the deep learning model learn an operation similar to the inverse 2D Fourier transformation parameterized by the complex $z_j$ matrix [11]. The function value $a_j(x, y)$ is computed by aggregating the powers of KxK (K = 32) frequencies, with each frequency's power being pondered by the complex $z_j$ matrix, which is defined for each output channel – and identical for all input channel locations. Thus, input channels close to each other are weighted in a similar manner by the $z_j$ matrix (especially for low values of $k$ and $l$ at low spatial frequencies). The attention scores are then obtained by applying the softmax function on $a_j(x, y)$ of all possible input channel locations. The value of each output channel $j$ can then be computed by

$$\forall j \in [D1], SA(X)^{(j)} = \frac{1}{\sum_{i=1}^{C} e^{a_j(x_i, y_i)}} (\sum_{i=1}^{C} e^{a_j(x_i, y_i)} X^{(i)}) \quad (5)$$

a weighted sum over all input channel locations. $C$ is the number of channels, and $X^{(i)}$ is the raw time series data for channel $i$. Note that since $z_j$ is learned with the classification task, the attention scores may only be relevant to the task at hand. In practical applications of spatial attention (SA), we normalize the coordinates $(x, y)$ by reducing the values to include a 0.1 margin on each side, as the variable $a_j$ is periodic. Additionally, for each output channel, we introduce spatial dropout by randomly selecting an input channel location $x_{drop}$, $y_{drop}$ and eliminating any sensor within a distance of 0.1 from that location in the softmax. This means that for each output channel, a random region of the 2-D channel space will be ignored. Following previous work [10], we applied a 1x1 convolution (filter size of 1) with no activation on the output of the spatial attention, using the same number of filters as the number of output channels.

**Channel harmonization using spatial attention.** Défossez and collaborators [10] used the spatial attention mechanism to take into account channel locations and potentially increase performance for a DL model trained on M/EEG data. We applied the spatial attention mechanism to perform channel harmonization by first noticing that the Fourier transform maps channel coordinates into a 2D frequency space supported by KxK frequencies. Thus, the matrix $z_j$ learns this mapping in a way that can handle all possible channel locations. This mapping is thus independent of the number of input channels and the location of these channels. The number of input channels then only affects the number of terms used by the softmax in equation (5). Thus, for each input EEG sample, we accordingly compute the $a_j$ values for each of its input channels given its location in equation (4), then the corresponding output channel values can be computed accordingly using equation (5).

### III. EXPERIMENTS

**Training, validation, and test sets.** While the dataset we used included 2,224 participants, there were only 787 females (35%). We decided to use 1574 participants (50% female) to ensure class balance by selecting the first 787 males from a list of participants ordered by their IDs. We checked the age distribution of the two subjects in the two classes matched [14]. Following [2], we then split the balanced data into training, validation, and test sets in a size ratio of 60:30:10. Each segment received a binary label, indicating a male (0) or a female (1). This led to 71,300 samples (885 participants; 49.94% female) for training, 39,868 samples (492 subjects; 50% female) for validation, and 16,006 samples (197 subjects; 50.3% female) for testing. There was no leakage of participants from the training set into the testing set.

**Experimental setup.** All models were trained on a single NVIDIA V100 SMX2 GPU (32 GB) with Python 3.7.10 and PyTorch 1.3.1 on the Expanse supercomputer. Generating the results in Table 2 and Table 3 required about 100 hours of GPU time. We trained all four models using an Adamax optimizer with default hyperparameters (learning rate = 0.002, $\beta_1 = 0.9$, $\beta_2 = 0.999$, $\epsilon = 1e-08$) except for decay set to 0.001. The batch size was set at 70 following [2]; training was performed for 15 epochs (see below). No other hyperparameter tuning nor batch normalization was performed.

**Data sampling and channel sub-sampling.** To generate a dataset with a different channel count, we sub-selected 23 channels from our original 128-channel data using the same channels as with previous works [2, 12]. Each sample in our dataset thus had dimensions of either 128x256 or 23x256 (number of channels and 2(s) x 128(Hz) time points). To generate the mix-channel data, we first performed data sub-sampling for each training, validation, and test set by splitting each set in half, ensuring balance between the two classes for each set and no overlapping of subjects across sets. For one half of each dataset, we performed channel sub-selection to retrieve a set of 23-channel data. The combination of 128-channel and 23-channel data for each training, validation, and test set constitutes the mixed-channel channel data for each of those sets. There was no leakage of participants between the 23-channel samples and the 128-channel samples.

**Validating spatial attention on individual datasets.** We set the baseline for our experiments by first training and evaluating the deep learning models without spatial attention on both 128-channel and 23-channel data individually (see Table 2). We then applied spatial attention and retrained the individual models to ensure that adding spatial attention would still allow the model to learn the gender classification task while not reducing the models' learning capacity on the data.

**Evaluating spatial attention performance.** The spatial attention mechanism allowed data trained on one set of channels (e.g., 128 channels) to be evaluated on data of the other set of channels (e.g., 23 channels) and hence also a combination of them (mixed dataset of both 128 and 23 channels). We also trained models on mixed-channel data (dataset with both 128-channel and 23-channel samples) and evaluated them on 128-channel, 23-channel, and mixed-channel data.

**Stopping and evaluation criteria.** Overfitting is a common issue in deep learning. One common DL practice to avoid overfitting is early stopping, in which training is stopped (and model performance evaluated) when validation accuracy (here, per-sample classification accuracy) starts to plateau or decrease as training accuracy continues to grow [7]. While training 128-channel data and 23-channel data models, we observed that the models converged and started to overfit the training data after about 15 epochs. Hence, for computational uniformity, we stopped the training of all subsequent models with all experiments after 15 epochs.

**Statistics.** To assess the robustness of our classification results, we trained each model 10 times using different random seeds, giving 10 different weight initializations, allowing us to calculate statistics using either unpaired 2-way ANOVA with variables *model* and *attention* (for Tables 2 and 3) or unpaired parametric t-test assuming unequal variances (for Table 3).

## IV. RESULTS

**128-channel model vs. 23-channel model with or without attention.** Table 2 shows that increasing the number of channels increases performance (average of 80.4% for the 128-channel model vs 78.0% for the 23-channel model; $F=40.7$; $p<3.10^{-7}$). Adding attention also increases performance (average of 83.7% with attention vs 80.4% without attention; $F=37.5$; $p<5.10^{-7}$). We did not notice an interaction between the two factors ($F=1.2$; ns).

| Model→<br>Attention↓ | 128-channel model | 23-channel model |
|---|---|---|
| No spatial attention | 80.4 (0.8) | 78.0 (1.8) |
| Spatial attention | 83.7 (1.5)✝ | 80.3 (1.4)✝ |

**Table 2.** Mean classification accuracy (and standard deviation in parenthesis) for the 23-channel and 128-channel models with or without using spatial attention. ✝ values are repeated in Table 3.

**Mixed-channel model overall performance.** Table 3 shows that the mided-channel model performed overall the best when considering all the different test data sets (mixed-channel model average performance of 80.3% vs. 70.3 for the 128-channel model ($F=142.6$; $p<10^{-12}$) and vs. 78.6% for the 23-channel model ($F=2.5$; $p=0.09$)). The difference with the 23-channel model is a trend at $p=0.09$ but would likely become significant with more repetitions given the significant difference observed when the two models are tested on 128-channel data – mean performance of 76.9% for the 23-channel model vs 81.4% for the mixed-channel model ($t=6$; $p=0.0001$).

| Model→<br>Test data↓ | 128 channel | 23 channel | Mixed channel |
|---|---|---|---|
| 128 channel | 83.7 (1.5) | 76.9 (1.2) | 81.4 (2.2) |
| 23 channel | 57.2 (3.0) | 80.3 (1.4) | 79.8 (2.5) |
| Mixed channel | 70.1 (4.2) | 78.6 (2.0) | 78.9 (1.6) |

**Table 3**. Classification accuracy (standard deviation in parenthesis) of models trained on different channel counts experiments, all while applying spatial attention.

**Mixed-channel model performance on homogenous test data.** We observed that models trained and tested on homogeneous 128-channel data give the best performance out of all models (83.7% vs. 81.4% on the same test data for the mixed model; $t=2.7$; $p=0.02$). However, the 128-channel and the mixed-channel models are the same models trained with different data – so this difference in performance does not reflect an issue with the architecture of the mixed-channel model. The model trained and tested on homogeneous 23-channel data did not differ significantly from the mix-channel model, though (80.3% vs 79.8% on the same 23-channel test data; $t=0.56$; ns).

**Models' performances on mixed-channel test data** (bottom row of Table 3)**.** Considering all models' performances on the mixed-channel test data, the models trained on 128-channel data performed significantly worse than the 23-channel and mixed model (t>5.8; p<0.0001 in both cases). The 23-channel and mixed models performed on par (78.6% vs 78.9% performance, respectively; t=0.36; ns). This could be an indication of data regularization by the 23-channel model, as the 23-channel data was subsampled from the 128-channel original dataset.

**Mixed-channel model vs 128-channel model performance on 23-channel test data.** The 128-channel model performed poorly on the 23-channel test data, with a mean performance of 57.2% vs 79.8% for the mixed-channel model (Table 3) ( t=18.5; p<$10^{-12}$). We interpret this poor result in the discussion.

**Mixed-channel model vs 23-channel model performance on 128-channel test data.** The 23-channel model performed worse than the mixed-channel model when evaluated on 128-channel test data, with a mean performance of 76.9% vs 81.4% (Table 3) (t=6; p=0.0001). However, we noticed that such performance is significantly better than the model trained on 128-channel data and evaluated on 23-channel data (76.9% vs. 57.2%, Table 3; t=19.3; p<$10^{-9}$). Since our 128-channel and 23-channel data originated from the same dataset originally, this indicates that the subsampling of 23-channel data acted as a form of regularization on the model.

## V. DISCUSSION

We have shown both spatial attention and the number of data channels improve the classification performance of DL models applied to EEG data.

We have also shown a mix-channel deep learning model using spatial attention outperformed models trained on a fixed 23 or 128-channel montage. This type of result is important because mixed-channel models may be trained with data from experiments using different channel montages.

Thus, the spatial attention mechanism can flexibly be used as a method to combine data with different channel counts for training deep learning models while taking into account the spatial information of the channels. Models trained on heterogeneous data can be effectively applied to homogeneous data of different channel counts without significant (if any) performance loss. Moreover, when applying such models on collections of data using different channel counts, models trained on mixed-channel data could potentially be highly performant. Channel harmonization using spatial attention thus gives a promising path forward for aggregating EEG data across datasets for the effective large-scale training of deep neural networks.

**Homogeneity of input channel locations.** For this work, while the number of input channels varies across training and testing sets, all data uses the same montage. This makes comparison easier, but it also has its drawbacks. The data regularization effect we observed when evaluating models trained on the subsampled 23-channel data on 128-channel and mixed-channel data might only hold for this experiment. We expect that in the situation where our data comes from heterogeneous sources with different channel montages, models trained on datasets of specific recording setup will only perform well when applied to data of the same recording parameters, further emphasizing the need for training deep models on non-uniform data samples. Thus, future work can apply the spatial attention mechanism to harmonize data from different datasets to test the efficacy of mixed-channel models.

**Generalization power of the 23 vs 128-channel model.** The 128-channel model failed to generalize to testing on 23-channel data. We think this could also be because the minimum distance between channels in the 23-channel data is large. This means that high frequencies (k or l equals 32) lead to more than 1 cycle with no smooth transition between neighbor channels. This pseudo-random information, multiplied by weights learned by the 128-channel model where higher spatial information was available, could lead to poor performance. However, this hypothesis would need to be tested with different channel montages.

In conclusion, spatial attention is useful to increase deep learning model performance and perform channel harmonization across datasets using different montages.

### CODE AND DATA AVAILABILITY

All data processing and model training code is made publicly accessible via Github: https://github.com/sccn/deep-channel-harmonization. We also made the raw data publicly available on Amazon Web Storage with full access instructions outlined in a previous publication [17].

### ACKNOWLEDGMENTS

Expanse supercomputer time was provided via XSEDE allocations and NSG (the Neuroscience Gateway). We thank Amitava Majumdar, Subhashini Sivagnanam, and Kenneth Yoshimoto for providing computational resources.